\pdfoutput=1

\documentclass[11pt]{article}
\usepackage{ACL2024}

\usepackage{times}
\usepackage{latexsym}

\usepackage{amsmath}
\usepackage{amsfonts}
\usepackage{graphicx} 
\usepackage{subfigure}
\usepackage{booktabs}
\usepackage{array}
\usepackage{enumitem}
\usepackage{amsthm}
\usepackage{algpseudocode}
\usepackage[switch]{lineno}
\usepackage[utf8]{inputenc}
\usepackage{eqparbox}
\usepackage[nopar]{lipsum}
\usepackage{multirow}
\usepackage{makecell}
\usepackage{xcolor}
\usepackage{hhline} 
\usepackage{microtype}
\usepackage{diagbox}
\usepackage{adjustbox}
\usepackage{multirow}

\usepackage{relsize}

\usepackage{booktabs, multirow, array}
\newcolumntype{N}{@{}m{0pt}@{}}

\usepackage[many]{tcolorbox}
\newtcolorbox{fancyquotes}{%
    enhanced jigsaw, 
    breakable,      
    frame hidden,   
    left=0.5cm,       
    right=0.1cm,      
    overlay={%
        \node [scale=8,
            text=black,
            inner sep=0pt,] at ([xshift=-1cm,yshift=-1cm]frame.north west){}; 
        \node [scale=8,
            text=black,
            inner sep=0pt,] at ([xshift=1cm]frame.south east){};  
            },
                parbox=false,
}

\usepackage{mathtools, nccmath}
\usepackage{scrextend}
\deffootnote[.25in]{.25in}{.15in}{\makebox[.25in][r]{\thefootnotemark .\hspace{.15in}}}

\makeatletter

\newtheorem*{proof*}{Proof}

\usepackage{algorithm}
\usepackage{listings}
\usepackage{color}

\definecolor{codeblue}{rgb}{0.25,0.5,0.5}
\def\@fnsymbol#1{\ensuremath{\ifcase#1\or \dagger\or *\or \ddagger\or
   \mathsection\or \mathparagraph\or \|\or **\or \dagger\dagger
   \or \ddagger\ddagger \else\@ctrerr\fi}}
   
\renewcommand{\arraystretch}{1.5}
\newcolumntype{C}[1]{>{\centering\let\newline\\\arraybackslash\hspace{0pt}}m{#1}}
\newcommand\ChangeRT[1]{\noalign{\hrule height #1}}

\ExplSyntaxOn
\NewExpandableDocumentCommand { \ValuePlusOne } { m } 
  { \int_eval:n { \int_use:c { c @ #1 } + 1 } }
\NewExpandableDocumentCommand { \Sec } { m } 
  { \fp_eval:n { secd ( #1 ) } }
\NewDocumentCommand { \Rot } { m }
  { 
    \hbox_to_wd:nn { 1 em }
      { 
        \hbox_overlap_right:n 
          { 
            \skip_horizontal:n { \fp_to_dim:n { 7 * cosd (\Angle) } } 
            \rotatebox{\Angle}{#1}
          } 
      } 
  }
\ExplSyntaxOff

\def\Angle{45}
    
\bigskip
\def\Angle{90}

\title{BGE Landmark Embedding: A Chunking-Free Embedding Method For Retrieval Augmented Long-Context Large Language Models} 

\author{
Luo Kun$^{1,2}$ \ \ \
Zheng Liu$^{1}$\thanks{Co-first and corresponding author} \ \ \
Shitao Xiao$^{1}$ \ \ \
Kang Liu$^{2}$ 
\\ 
1: Beijing Academy of Artificial Intelligence \ \ \ 
2: Chinese Academy of Sciences \\
{\tt 42002074@xs.ustb.edu.cn} \ \ \ 
{\tt zhengliu1026@gmail.com} 
}

\begin{document}
\maketitle 

\begin{abstract} 
Retrieval augmentation is a promising approach to handle long-context language modeling. However, the existing retrieval methods usually work with the chunked context, which is prone to inferior quality of semantic representation and incomplete retrieval of useful information. In this work, we propose a new method for the retrieval augmentation of long-context language modeling, called \textbf{Landmark Embedding}. Our method is characterized by threefold technical contributions. Firstly, we introduce a \textit{chunking-free architecture}, which keeps the long context coherent such that high-quality embeddings can be generated for the fine-grained units within the context. 
Secondly, we present a \textit{position-aware objective function}, which prioritizes the ultimate boundary for a consecutive span of information. 
By learning to discriminate such a special position, the useful information can be comprehensively retrieved for the query.  
Thirdly, we design a novel \textit{multi-stage learning algorithm}, which makes the best use of readily available data and synthetic data for cost-effective training of the landmark embedding. In our experimental study, landmark embedding is able to substantially improve the performance for both LLaMA-2 and ChatGPT in a variety of long-context tasks; meanwhile, it also outperforms the existing retrieval methods with a notable advantage. Our model and code will be made publicly available\footnote{https://github.com/FlagOpen/FlagEmbedding}.

\end{abstract}

\begin{figure}[t]
\centering
\includegraphics[width=1.0\linewidth]{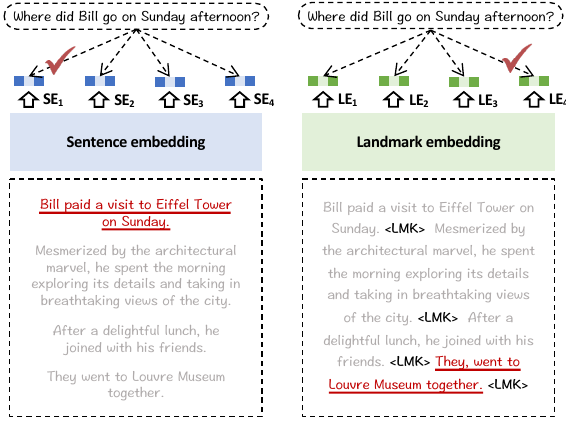}
\vspace{-20pt}
\caption{\textbf{Sentence Embedding} works with the chunked context, which tends to select the salient sentence. \textbf{Landmark Embedding} maintains a coherent context, which enables it to select the right sentence.} 
\vspace{-15pt}
\label{fig:1}
\end{figure}

\vspace{-5pt} 
\section{Introduction}
Large language models (LLMs) need to handle long-sequence inputs when dealing with many important applications, such as question answering and reading comprehension \cite{bai2023longbench}. Unfortunately, the existing LLMs are usually constrained by a limited size of context window, e.g., 2K for LLaMA-1 \cite{touvron2023llama-a} and 4K for LLaMA-2 \cite{touvron2023llama-b}. Although the size of context window can be extended through fine-tuning over long-sequence data \cite{chen2023longlora,longchat2023,peng2023yarn}, the fine-tuned model could incur a considerable cost in both training and inference, and exerts an unfavorable impact to LLMs' original capabilities. Recently, the retrieval-augmentation emerges as a promising option to facilitate long-context language modeling \cite{xu2023retrieval_meets,bai2023longbench,zhang2023retrieve}. It employs a standalone retriever where useful information can be filtered and presented as a concise input. The above working mechanism is simple, efficient, and well-compatible with the downstream LLMs. 

With a long-sequence input, the typical retrieval augmentation workflow is performed with three steps: 1) chunking, 2) embedding, and 3) retrieval. In the first place, it partitions the long-sequence input into a list of chunks. Then, it encodes each chunk into its embedding. Finally, it retrieves the useful chunks for the query based on the embedding similarity. The chunking strategy is a very tricky problem in practice. As widely discussed by many popular RAG frameworks, like Langchain \cite{langchain}, LlamaIndex \cite{llamaindex}, Pincone \cite{pincone}, this problem is usually tackled by empirical or heuristic methods. However, no matter what chunking strategy is used, two inherent limitations are inevitable. On one hand, the input sequence is partitioned into disconnected chunks. Consequently, it will break the coherence of context which is unfavorable to the quality of embedding. On the other hand, it is also likely to split the consecutive information into different chunks. The salient chunks can be easily retrieved; nevertheless, other useful but less salient chunks can be overlooked, which results in the incomplete retrieval of necessary information. 

In this paper, we come up with the \textbf{Landmark Embedding}, a new embedding method optimized for the retrieval-augmentation of long-context language modeling. The new method is highlighted by its technical contributions in three perspectives. 

Firstly, we introduce a \textit{chunking-free model architecture}, where embeddings for the fine-grained input units, e.g., sentences, can be generated based on a coherent long context. The new architecture employs a group of special tokens, namely the landmarks (LMK), and dispatches them to the end of each sentence. At the same time, it takes advantage of a LLM-based encoder to jointly process the landmarked long context. Thanks to the perception of rich contextual information, the landmark embedding can be a highly discriminative representation of each sentence, which presents a critical improvement over the  conventional sentence embeddings generated from a chunked context. Besides, the new architecture will resort to a sliding window, where landmark embeddings can be generated for an arbitrary long context via stream processing. 

Secondly, we propose a \textit{position-aware objective function} to facilitate the complete retrieval of useful information. As discussed, one piece of information tends to be jointly conveyed by multiple consecutive sentences within the long context. Instead of treating them equally as positive positive samples, we assign each sentence with a differentiated weight which grows exponentially with its position in the context. As a result, the last sentence, i.e. the ultimate boundary of the information, will be emphasized and better discriminated. With the jointly selection of the front-$k$ sentences before the ultimate boundary, the useful information to the query can be comprehensively included. 

Thirdly, we design a \textit{multi-stage learning algorithm}, where different training strategies and data sources can be jointly used to facilitate the training of landmark embedding. The typical embedding model is trained by paired texts, e.g., question answering; however, such data is not directly suitable for our scenario. To address this problem, the new algorithm factorizes landmark embedding with two basic capabilities: the fundamental semantic discriminability, and the high-level contextualized representation capability, which can be progressively established in three steps: 1) \textit{distant supervision} over pairwise data, 2) \textit{weak supervision} over noisy long-context data generated by rules, 3) \textit{fine-tuning} over high-quality long-context data synthesized by LLMs. The above workflow can make the best use of readily available data (adequate but less relevant) and synthetic data (relevant but inadequate), which leads to a superior cost-effectiveness of training. 


We empirically analyze landmark embedding based on 2 popular LLMs: LLaMA-2-7B (chat) with a short context window (4K), ChatGPT-3.5 (turbo) with a much longer context window (16K). The experiment is performed on top of 6 long-context evaluation datasets. Most of the evaluation samples are far beyond the coverage of the 4K context window, while a large portion of them are within the 16K context window. In our experiment, landmark embedding achieves a substantial advantage over both the LLaMA-2-7B baseline and the retrieval-augmentation results powered by the existing retrieval methods. Meanwhile, it also notably improves the performance of ChatGPT-3.5 using a much shorter input context. Such a result overturns the previous conclusion that retrieval-augmentation can only benefit the LLMs of weak long-context capabilities \cite{bai2023longbench}, which indicates a more extensive usage of the corresponding techniques. 

To summarize, the following contributions are made in this work. 1) We propose landmark embedding. To the best of our knowledge, it is the first embedding model which performs systemic optimization for the retrieval augmented long-context language modeling. 2) Our method presents three technical advantages: the chunking-free model architecture, the position-aware objective function, and the multi-stage learning algorithm, which jointly contribute to the superior capability of our embedding model. 3) We perform comprehensive experiments with LLaMA-2 and ChatGPT, whose result verifies the effectiveness of landmark embedding, and indicates a broader application scope of retrieval techniques in dealing with the long-context tasks.

\section{Related Work}
The related works are reviewed from three aspects: long-context language modeling, retrieval-augmentation, embedding-based retrieval methods. 

First of all, a large body of research works have been dedicated to the extension of LLMs' lengths from different directions. One common practice is to modify the position encoding mechanism, where the LLMs trained on short texts can directly handle longer input during the inference time \cite{chen2023extending,ntkaware2023}. Despite simplicity, such methods are prone to inferior performances without further fine-tuning. Another popular method is to take advantage of continual training, where the existing LLMs are fine-tuned over long-sequence data to establish a longer context window \cite{longchat2023,chen2023longlora,peng2023yarn,tworkowski2023focused,mohtashami2023landmark}. 
However, the fine-tuning based methods are prone to two subsequent problems. On one hand, the fine-tuned LLM will incur an expensive cost for both training and inference. On the other hand, the fine-tuning operation could be unfavorable to the LLM's performance with short-sequence inputs. Apart from the above common approaches, the LLM's context can also be extended by context compression \cite{chevalier2023autocompressors,zhang2024soaring} and stream processing \cite{xiao2023efficient,han2023lm_infinite}. Nevertheless, the compression methods are likely to result in information loss, while the stream processing will discard the useful information beyond the sliding window. It remains to explore more effective methods for long-context language modeling in the future.

In the meantime, the retrieval-augmented generation (RAG) is another important issue for LLMs' research. Typically, it employs a standalone retriever, where useful information can be introduced from a vast open-world corpus to enhance the LLM's generation quality \cite{lewis2020retrieval,guu2020retrieval,borgeaud2022improving}. Previously, RAG used to be applied for knowledge-intensive tasks, such as open-domain question answering and fact verification \cite{petroni2020kilt}, where an external knowledge base is presented. Recently, retrieval augmentation is also found helpful to long-context language modeling, as useful information can be retrieved and presented as a concise input for the LLM \cite{xu2023retrieval_meets,bai2023longbench,zhang2023llmembedder}. Compared with other alternative methods on context extension, the retrieval-based methods are distinguished for the simplicity and compatibility, as they don't need modification of the downstream LLM, and can be easily combined with other methods to establish a longer context. 

Finally, the RAG system usually works with an embedding model to retrieve the useful information. In the past few years, many critical techniques have been well established for the effective learning of embedding models, e.g., pre-training \cite{gao2021condenser,xiao-etal-2022-retromae,xiao2023retromaev2,wang2022simlm,li2023making}, hard-negative sampling \cite{xiong2020approximate,ren2021rocketqav2}, knowledge distillation \cite{hofstatter2020improving,chen2024bge}, etc. On top of these techniques, there have been a number of powerful embedding models developed for the general-purpose retrieval applications \cite{izacard2021unsupervised,ni2021sentence,neelakantan2022text,wang2022text,xiao2023c}. However, the existing methods rely on chunking when dealing with the retrieval augmentation of long-context language modeling. As a result, it will inevitably break the coherence of context, which is prone to inferior quality of embedding and incomplete retrieval of useful information. 





\section{Landmark Embedding}
\subsection{Preliminary}
The LLM presents a unified foundation to solve arbitrary NLP tasks through language modeling. Given the input prompt, the LLM optimizes the generation likelihood of the target answer ($X$) in the form of auto-regression. For a wide variety of applications, e.g., question answering and reading comprehension, the input prompt can be explicitly split into context ($ctx$) and query ($q$). Without loss of generality, the LLM's generation objective can be presented as the following function:  
\begin{equation}
    max. \log \mathrm{LLM}(x_t|q, ctx, X_{<t}). 
\end{equation}
In many situations, the input context is too long to fit into existing LLM's context window. To address this problem, the retrieval-based method seeks to compress the context by selecting the most useful parts from it. Typically, it will chunk the context into: $S: \{s_1, ..., s_N\} \leftarrow chunk(ctx)$, and select the top-$k$ chunks based on a retrieval model $\mathrm{\gamma(\cdot)}$: 
\begin{equation}
    S^*: \{s_1, ..., s_k\} \leftarrow \textit{top-}k. \{s: \gamma(q,s) | S\}. 
\end{equation}
One critical step for the above workflow is chunking. As introduced, the chunking operation is very tricky, which needs to be conducted empirically or heuristically. It will always break the coherence of context, leading to an inferior embedding quality and a higher probability of incomplete retrieval. In this work, we target on a new retrieval method $\gamma'(\cdot)$ without the dependency on chunking operation. Notably, it will let the useful information to be directly retrieved from a coherent context: 
\begin{equation}
    C^*: \{c_1, ..., c_k\} \leftarrow \gamma'(q, ctx). 
\end{equation}
In this place, $c_i$ indicates a fine-grained unit of the input context, e.g., a sentence. With the perception of contextual information, the underlying semantics about each fine-grained unit can be effectively represented, which facilitates the accurate retrieval of relevant information for the query.

\subsection{Chunking-Free Architecture} 
\begin{figure}[t]
\centering
\includegraphics[width=1.0\linewidth]{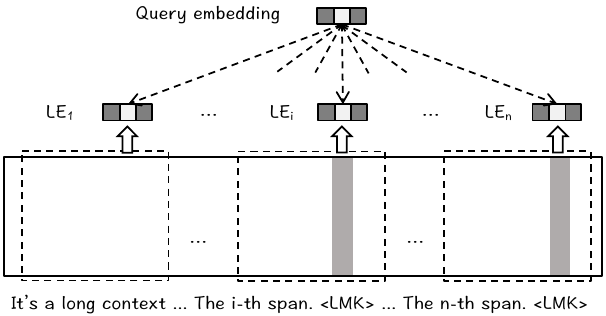}
\caption{\textbf{Architecture for Landmark Embedding}. The landmark LMK token is appended to the end of each sentence. A sliding window is employed to handle the input sequence longer than the LLM's context window.}  
\label{fig:2}
\end{figure}

We propose a novel embedding model, whose architecture is shown in Figure \ref{fig:2}. Suppose the input context is composed of $n$ sentences: $ctx: \{c_1, ..., c_n\}$. Instead of chunking the input context into disconnected segments, it dispatches a special token, called the landmark (LMK), to the end of each sentence. The landmark is used to capture the underlying semantics for its corresponding sentence. Particularly, the landmark is jointly encoded with the sentence and neighboring context, where the output embedding, a.k.a. the landmark embedding (LE), is utilized for representation of the sentence. In our work, we take advantage of a large language model (e.g., LLaMA-2-7B) as the encoding backbone, which brings forth two benefits: 1) it substantially contributes to the quality of representation thanks to the LLM's superior expressiveness, 2) it can incorporate adequate neighboring context based on the LLM's long context window. The same encoder is also utilized for the generation of query's embedding. Formally, the generation of landmark embedding and query embedding are presented by the following functions: 

\begin{align*}
    & \mathrm{LE}_i \leftarrow \mathrm{LLM}(c1,...,c_i; \mathrm{LMK}).embed[-1], \\
    & \mathrm{E}_q \leftarrow \mathrm{LLM}(query; \mathrm{LMK}).embed[-1]. 
\end{align*}
Based on the above result, the relevance between the query and each sentence is computed as the inner product of the two embeddings: $\langle \mathrm{E}_q, \mathrm{LE}_i \rangle$.

Note that the input context can be even longer than the LLM's context window. To handle this problem, we leverage a sliding window where the long context can be streamingly processed. In this situation, the generation of landmark embedding will be conducted as: 
\begin{equation*}
    \mathrm{LE}_i \leftarrow \mathrm{LLM}(c_{i-l},...,c_i; \mathrm{LMK}).embed[-1], 
\end{equation*}
where $l$ indicates the number of sentences within the current sliding window.

\subsection{Position-Aware Objective}\label{sec:position}
The landmark embedding is learned by contrastive learning, where the query and its relevant sentences can be distinguished by the higher embedding similarities. The useful information to the query tends to gather as multiple consecutive sentences within the context: $\{c_{z-m}, ..., c_{z}\}$. As a result, we can derive the following general form of loss function for the contrastive learning: 
\begin{equation}\label{eq:basic_obj}
    min. - \sum_q \sum_{i \leq m} \lg \frac{\exp(\langle \mathrm{E}_q, \mathrm{LE}_{z-i}\rangle)}{\sum_{j=1...n} \exp(\langle\mathrm{E}_q, \mathrm{LE}_j\rangle)}. 
\end{equation}
With the above formulation of loss function, the landmark of each relevant sentence is assigned with a positive label of equal importance. Nevertheless, the basic formulation is problematic knowing that it may let the most salient sentence (e.g., the one with the most overlapping keywords with the query) get the highest similarity. In our work, we target on the complete retrieval of useful information. Therefore, we make an emphasis on the ultimate boundary where the whole consecutive sentences can be comprehensively included. Although it may simply assign the last landmark with a positive label, we propose to leverage all landmarks because of their valid relevance with the query. Particularly, we differentiate their importance by introducing the positional weight $w_i$ for sentence $c_{z-i}$: $w_{i} \leftarrow \exp(-\alpha*i)$, where $\alpha$ is the temperature parameter. Based on the positional weight, we modify the basic contrastive learning with the position-aware objective function: 
\begin{equation}\label{eq:weight_obj}
    min. - \sum_q \sum_{i \leq m} \lg \frac{w_i*\exp(\langle \mathrm{E}_q, \mathrm{LE}_{z-i}\rangle)}{\sum_{j=1...n} \exp(\langle\mathrm{E}_q, \mathrm{LE}_j\rangle)}. 
\end{equation}
The position-aware objective presents two benefits: 1) the relevant sentences can be fully utilized for the training of landmark embedding, 2) the ultimate boundary of the useful information can be emphasized and better discriminated.

\subsection{Multi-Stage Learning}\label{sec:multi}



The typical training data of embedding model consists of paired texts, e.g., question and answer, which is seemingly inappropriate for the training objective in Eq \ref{eq:weight_obj}. However, we argue that the functionality of landmark embedding can be factorized with two fundamental capabilities: 1) the basic semantic discriminability, 2) the contextualized representation capability, i.e., representing each sentence w.r.t. its context. Based on this argument, we design the multi-stage learning algorithm, which enables the two capabilities to be progressively established on top of proper training data. In the first place, the landmark embedding is initialized as a general sentence-level embedding model. Afterwards, it is enhanced as a contextual representation model where discriminative embeddings can be generated for its included sentences. The progressive training takes place with three steps.

$\bullet$ \textbf{Distant supervision}. Firstly, we make use of the pairwise training data from MS MARCO \cite{nguyen2016ms}, based on which the model can be initialized as a basic sentence embedder. In this place, the landmark embedding takes a special form as only one single landmark is appended to the end of answer's context: $\mathrm{LE}_{a} \leftarrow \mathrm{LLM}(answer; \mathrm{LMK}).embed[-1]$. The first-stage training follows the basic training form of dense retrieval, where 15 hard negatives together with the in-batch negatives are presented for each query. 

\begin{figure}[t]
\centering
\includegraphics[width=1.0\linewidth]{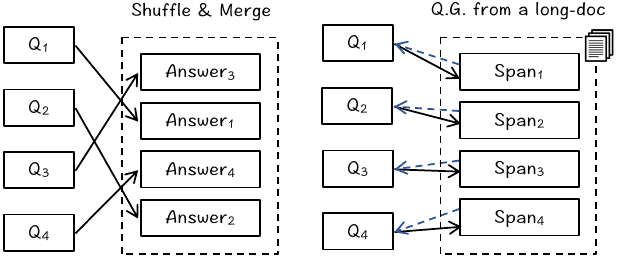}
\vspace{-10pt}
\caption{\textbf{Weak Supervision (L) and Fine-Tuning (R)}. }  
\label{fig:3}
\end{figure}

$\bullet$ \textbf{Weak Supervision}. In the second step, we make a simple modification of the pairwise training data where the model can be trained to generate discriminative sentence embeddings within a long context. Particularly, we randomly shuffle the answers from different queries, and merge them as one pseudo long document (left half of Figure \ref{fig:3}). Therefore, the embedding for the $i$-th answer can be generated as: $\mathrm{LE}_{a_i} \leftarrow \mathrm{LLM}(a_{j{\neq}i}, ..., a_i; \mathrm{LMK}).embed[-1]$. The second stage still relies on in-batch negatives, where the landmark embeddings from other answers $\mathrm{LE}_{a_{j \neq i}}$ are utilized as the negative samples.

\begin{table}[t]
    \centering
    \footnotesize
    \setlength{\tabcolsep}{2pt}
    \begin{tabular}{l|C{1.0cm}C{1.0cm}C{1.0cm}C{1.0cm}|C{1.0cm}}
        \toprule
        & $\leq$4K & $\leq$8K & $\leq$12K & $\leq$16K & Total \\ 
        \hline
        Stage II.  & --  & --  & --  & 240K & 240K  \\
        Stage III. & 40K & 30K & 10K & 10K  & 90K \\
        \bottomrule
    \end{tabular}
    \caption{\textbf{Distribution of training data's lengths.}}
    \label{tab:train-data}
    \vspace{-10pt}
\end{table}

$\bullet$ \textbf{Fine-Tuning}. We leverage synthetic data for the final stage of fine-tuning. In this step, we make use of the real-world long documents from Wikipedia\cite{wikidump}. For each long-document, a series of text spans are randomly sampled, where pseudo queries are generated by prompting the LLM\footnote{We make use of ChatGPT-35-turbo's API in this work: \href{https://openai.com/blog/chatgpt}{https://openai.com/blog/chatgpt}}.  The synthesized data will incur an extra cost due to the calling of LLM API. Besides, it may also be distinct from the real-world data distribution. Therefore, only a small amount of synthetic data is generated for the final training stage. However, thanks to the fundamental capabilities established in the first two stages, landmark embedding can achieve a superior performance after moderate fine-tuning. Detailed information of training data and curating method is shown in Appendix~\ref{syn data}

\section{Experiment}
The experimental study focuses on the following three research questions. \textbf{RQ 1.} The exploration of landmark embedding's impact on the retrieval augmentation of long-context language modeling. \textbf{RQ 2.} The comparison between landmark embedding and the existing retrieval methods based on chunked contexts. \textbf{RQ 3}. The analysis of technical factors in landmark embedding. 

\begin{figure}[t]
\centering
\includegraphics[width=0.97\linewidth]{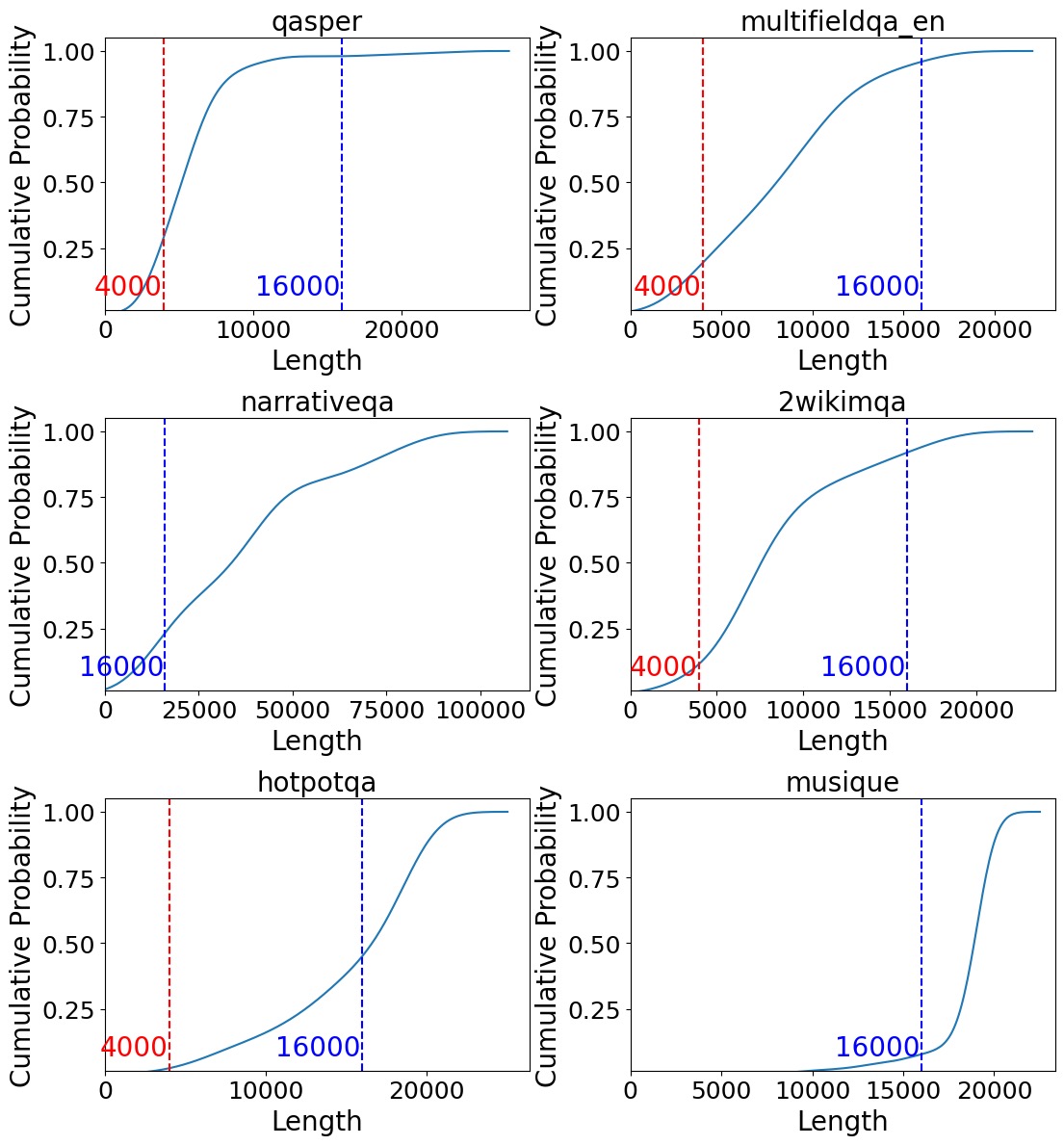}
\vspace{-5pt}
\caption{\textbf{Length distribution of evaluation data}. }  
\vspace{-10pt}
\label{fig:4}
\end{figure} 

\renewcommand{\arraystretch}{1.2}
\begin{table*}[t]
    \centering
    \footnotesize
    \setlength{\tabcolsep}{6pt}
    \begin{tabular}{c|c|c|c|cccccc|c}
    \ChangeRT{1pt} 
    LLM & Retriever & Unit & Len. & NQA & QASP & MFQA & HQA & 2WIKI & MSQ & Avg. \\
    \hline
    \multirow{9}{*}{Llama2-7B-chat} & w/o retrieval & - & 3,500 & 18.7 & 19.2 & 36.8 & 25.4 & 32.8 & 9.4 & 23.7 \\
     & Contriever & chunk & 2,275 & 18.3 & 23.8 & 41.8 & 33.6 & 34.5 & 17.2 & 28.2 \\
     & OpenAI-2 & chunk & 2,275 & 20.0 & 25.7 & 40.3 & 34.7 & 34.4 & 17.3 & 28.7 \\
     & BGE-large & chunk & 2,275 & 17.6 & 21.7 & 45.4 & 34.3 & \textbf{36.9} & 19.9 & 29.3 \\
     & E5$_{\mathrm{\text{mistral-7b}}}$ & chunk & 2,275 & \textbf{21.6} & 24.1 & 42.2 & 37.6 & 31.4 & 20.7 & 29.6 \\
     & Contriever & sentence & 2,190 & 16.2 & 26.5 & 44.4 & 33.5 & 33.3 & 17.5 & 28.6 \\
     & BGE-large & sentence & 2,190 & 17.9 & 24.4 & 46.3 & 37.4 & 35.0 & 21.3 & 30.3 \\
     & E5$_{\mathrm{\text{mistral-7b}}}$ & sentence & 2,190 & 16.5 & 24.0 & 47.3 & 37.6 & 35.4 & \textbf{21.7} & 30.4 \\
     \cline{2-11} 
     & Ours & sentence & 2,190 & 21.3 & \textbf{27.7} & \textbf{47.6} & \textbf{40.2} & 36.3 & \textbf{21.7} & \textbf{32.5} \\
     \hline
    \multirow{9}{*}{ChatGPT-3.5-turbo} & w/o retrieval & - & 15,500 & \textbf{23.6} & \textbf{43.3} & 52.3 & 51.6 & 37.7 & 26.9 & 39.2 \\
     & Contriever & chunk & 2,275 & 18.3 & 35.6 & 54.3 & 47.0 & 39.5 & 25.2 & 36.6 \\
     & OpenAI-2 & chunk & 2,275 & 21.8 & 38.1 & 52.8 & 46.6 & 44.9 & 30.4 & 39.1 \\
     & BGE-large & chunk & 2,275 & 21.9 & 37.2 & 49.1 & 49.5 & 42.2 & 30.4 & 38.4 \\
     & E5$_{\mathrm{\text{mistral-7b}}}$ & chunk & 2,275 & 21.0 & 41.2 & 49.2 & 54.0 & 43.7 & 27.2 & 39.4 \\
     & Contriever & sentence & 2,190& 17.5 & 41.0 & 50.2 & 46.2 & 41.9 & 24.1 & 36.8 \\
     & BGE-large & sentence & 2,190 & 19.8 & 41.2 & 51.3 & 50.5 & \textbf{46.5} & 29.6 & 39.8 \\
     & E5$_{\mathrm{\text{mistral-7b}}}$ & sentence & 2,190 & 20.0 & 39.0 & 49.4 & 55.4 & 45.9 & \textbf{31.1} & 40.1 \\
     \cline{2-11} 
     & Ours & sentence & 2,190 & 22.3 & 42.7 & \textbf{55.7} & \textbf{56.1} & 46.2 & 29.5 & \textbf{42.1} \\
    \ChangeRT{1pt} 
    \end{tabular}
    \vspace{-5pt}
    \caption{Experiment results on retrieval augmented long-context language modeling. “unit” denotes chunking and evidence selecting method. “Len.” denotes the average token number for the answering model(LLM).}
    \vspace{-10pt}
    \label{tab:5}
\end{table*}

\subsection{Settings}
We utilize two popular LLMs for RAG in our experiment. One is the \textbf{LLaMA-2-7B} (chat) model. It is a lightweight open-source LLM, whose context length is \textbf{4K}. The other one is the \textbf{ChatGPT-3.5} (turbo). It is a more powerful but closed-source LLM, whose context length is \textbf{16K}. The evaluations are performed with the following long-context language understanding datasets from {LongBench} \cite{bai2023longbench}, where explicit queries are available for the evaluation samples: \textbf{NarrativeQA} \cite{kovcisky2018narrativeqa}, \textbf{Qasper} \cite{dasigi2021dataset}, \textbf{MultifieldQA} \cite{bai2023longbench}, \textbf{HotpotQA} \cite{yang2018hotpotqa}, \textbf{2WikiMQA} \cite{ho2020constructing}, \textbf{MuSiQue} \cite{trivedi2022musique}. The first three datasets are about single-doc QA where the useful information is concentrated in the long context. The last three datasets are about multi-doc QA where useful information may exist in different parts of the long context. We follow LongBench\cite{bai2023longbench} using F1 score as the evaluation metric. It is worth noting that the above datasets are differentiated in their sequence lengths. As demonstrated by Figure \ref{fig:4}, the majority of evaluation samples are longer than 4K, which is far beyond the context length of LLaMA-2. However, many of them are shorter than 16K, especially for Qasper, MultifieldQA, 2WikiMQA, and HotpotQA, which is within the coverage of ChatGPT-3.5-turbo. 

We consider the following \textbf{baseline methods}. 1) Contriever \cite{izacard2021unsupervised}, 2) OpenAI Text Embedding (Ada-002) \cite{neelakantan2022text}, 3) BGE-v1.5-large  \cite{xiao2023c}, 4) E5-Mistral \cite{wang2023improving}. Notably, E5-Mistral is the state-of-the-art text embedding model upon the time of this paper. It is trained from a Mistral-7B model \cite{jiang2023mistral}, which achieves the leading performance on MTEB benchmark \cite{MTEB} with an overwhelming advantage. The baseline retrievers utilize two alternative chunking strategies. One is chunking by \textbf{sentences}; the other one is chunking by equal-sized \textbf{text spans}. In our work, each text span is made up of 200 words as empirically determined by Longbench \cite{bai2023longbench}. The baselines will select the top-N chunks for each query, and will take their front and back sentences together as evidence for retrieval augmentation. We select top-7 chunks for span-based chunking and top-15 chunks for sentence-based chunking, which leads to similar context lengths. As landmark embedding is to identify the ultimate boundary of information, it will retrieve the font two sentences together with its top-N results. 

Landmark embedding is based on a \textbf{LLaMA-2-7B} backbone \cite{touvron2023llama-b}, whose context is extended to \textbf{32K by LongLora} \cite{chen2023longlora}. All training operations take place on a single 8×A100 (40GB) GPUs. The learning rate is 1×$10^{-4}$, the weight decay is 1×$10^{-6}$. The batch size for the 1st-stage training is 32; the batch size for the 2nd and 3rd stage training is 1, where we accumulate the gradient over 64 steps. We leverage Flash-attention-v2 \cite{dao2023flashattention}, Gradient Checkpointing \cite{chen2016training}, and Deepspeed-Zero \cite{rajbhandari2020zero} to speed up the training. 

\begin{table}[t]
    \centering
    \footnotesize
    \setlength{\tabcolsep}{2pt}
    \begin{tabular}{p{1.0cm}|C{1.2cm}|C{1.7cm}|C{1.4cm}C{1.5cm}}
        \ChangeRT{1pt} 
        Dataset & Doc Len. & Method & MRR@10 & Recall@10 \\
        \hline
        \multirow{4}{*}{Wiki} & \multirow{4}{*}{6,748} & Contriever & 79.74 & 96.11 \\ 
        & & BGE-large & 88.32 & 98.70 \\ 
        & & E5$_{\mathrm{\text{mistral-7b}}}$ & 91.42 & 99.01 \\ 
        \cline{3-5}
        & & Ours & \textbf{95.21} & \textbf{99.60} \\ 
        \hline
        \multirow{4}{*}{Arxiv} & \multirow{4}{*}{9,982} & Contriever & 66.27 & 94.12 \\ 
        & & BGE-large & 78.82 & 97.06 \\ 
        & & E5$_{\mathrm{\text{mistral-7b}}}$ & 81.37 & 97.65 \\ 
        \cline{3-5} 
        & & Ours & \textbf{84.72} & \textbf{98.43} \\ 
        \ChangeRT{1pt} 
    \end{tabular}
    \vspace{-5pt}
    \caption{\textbf{Pilot experiment on retrieval accuracy.}}
    \vspace{-10pt}
    \label{tab:6}
\end{table}

\subsection{Main Result}
The experiment result on retrieval augmented long-context language modeling is presented in Table \ref{tab:5}, where the following observations can be made.  

\subsubsection{Analysis on retrieval augmentation}
Our method achieves a remarkable retrieval augmentation effect, as it consistently {outperforms the basic LLaMa-2-7B}, i.e., w/o retrieval, in every evaluation task, which ultimately results in a remarkable improvement of +8.8 points in terms of the average performance. At the same time, our method also brings forth the biggest improvement in comparison with other baseline retrievers. 

The retrieval-augmentation's impact is relatively smaller with ChatGPT-3.5, as most of the baseline retrievers are unable to improve the performance of w/o retrieval. Such an observation is consistent with the reported result in recent study \cite{xu2023retrieval_meets}, and it is intuitive to understand this result considering that the context length of ChatGPT-3.5 is expanded to 16K. With such a large context window, ChatGPT can intake more than 15K input tokens for each evaluation sample, whereas the retrieval augmentation methods only utilize about 2K input tokens. In many situations, the evaluation samples can almost be fully covered by such a long context window (Figure \ref{fig:4}), which means the retrieval methods can hardly introduce extra information outside ChatGPT's context.

\begin{figure}[t]
\centering
\includegraphics[width=0.95\linewidth]{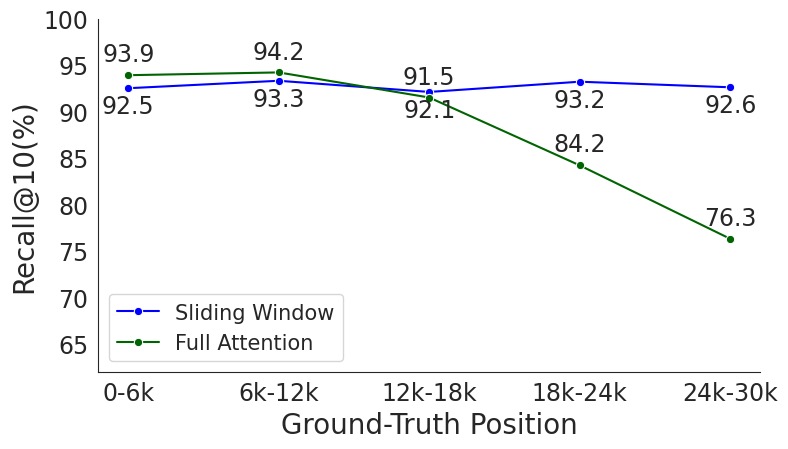}
\vspace{-10pt}
\caption{\textbf{Needle in a haystack test}. }  
\vspace{-15pt}
\label{fig:5}
\end{figure}

\begin{table*}[t]
    \centering
    \footnotesize
    \setlength{\tabcolsep}{6pt}
    \begin{tabular}{c|c|c|C{0.9cm}C{1.0cm}C{1.0cm}C{0.9cm}C{1.0cm}C{0.9cm}|C{0.9cm}}
    \ChangeRT{1pt} 
    Train Objective & Retrieval & Unit & NQA & QASP & MFQA & HQA & 2WIKI & MSQ & Avg. \\
    \hline
    w/o Position-aware & Surround-$k$ & sentence & 19.1 & 29.8 & 46.8 & \textbf{40.2} & 34.2 & 17.0 & 31.2 \\
    w/o Position-aware & Front-$k$  & sentence & 19.4 & 28.5 & 46.5 & 38.5 & 33.8 & 19.0 & 31.0 \\
    w. Position-aware   & Surround-$k$ & sentence & 19.4 & \textbf{29.7} & \textbf{47.9} & 39.0 & 36.0 & 17.8 & 31.6 \\
    w. Position-aware*   & Front-$k$  & sentence & \textbf{21.3} & 27.7 & 47.6 & \textbf{40.2} & \textbf{36.3} & \textbf{21.7} & \textbf{32.5} \\ 
    \hline\hline
    Stage I. only    & Front-$k$  & sentence & 18.9 & 27.0 & 45.0 & 35.5 & 33.2 & 17.2 & 29.4 \\
    Stage II. only   & Front-$k$  & sentence & 19.0 & 27.4 & 43.9 & 34.4 & 32.7 & 16.5 & 29.0 \\
    Stage III. only  & Front-$k$  & sentence & 20.5 & 27.2 & 45.3 & 39.2 & 34.3 & 15.3 & 30.3 \\
    Stage I. + II.   & Front-$k$  & sentence & 19.2 & 26.5 & 47.0 & 36.2 & 32.8 & 16.8 & 29.8 \\
    Stage II. + III. & Front-$k$  & sentence & 19.4 & 26.7 & 46.8 & 39.8 & 35.4 & 18.3 & 31.0 \\
    All three stages*  & Front-$k$  & sentence & \textbf{21.3} & \textbf{27.7} & \textbf{47.6} & \textbf{40.2} & \textbf{36.3} & \textbf{21.7} & \textbf{32.5} \\
    \ChangeRT{1pt} 
    \end{tabular}
    \vspace{-5pt}
    \caption{\textbf{Ablation study.} \textbf{Upper}: impact from position-aware objective. \textbf{Lower}: impact from multi-stage learning.}
    \vspace{-15pt}
    \label{tab:8}
\end{table*}

Despite these challenges, our method can still outperform ChatGPT-3.5, which leads to a +2.9 points improvement in the average performance. It consistently outperforms ChatGPT in the multi-doc QA tasks, i.e. HQA, 2WIKI, MSQ; meanwhile, it achieves improved or comparable performances in the single-doc QA tasks. The distinction between the two tasks is probably because the useful information tends to be more scattered and exists within multiple documents in the multi-doc scenario, while it is more concentrated in the single-doc scenario. It is also worth noting that our method only works with 2,190 input tokens, which is much less than the 15,500 tokens used by ChatGPT. In other words, its empirical advantage is achieved along with a higher running efficiency. 

\subsubsection{Pilot analysis on retrieval}
In addition to the end-to-end performance on the above long-context tasks, we conduct pilot experiments for more detailed analysis about the retrieval accuracy. In Table \ref{tab:6}, we leverage the hold-back test set of the synthetic data from Wikipedia for evaluation (1000 samples in total). We also curate the synthetic testing samples based on ArXiv\cite{clement2019arxiv} documents (500 samples in total), which will reflect the retriever's generalization with the o.o.d. corpus. For both datasets, our method can achieve a much higher retrieval accuracy than the baseline retrievers which rely on the chunked context. Besides, we also perform the needle in a haystack test as Figure \ref{fig:5}, where the ground-truth document span is randomly placed in 30K context \cite{liu2023lost,ivgi2023efficient}. Detailed setting is described in Appendix~\ref{needle in a haystack test}. We compare two alternative formulations of landmark embedding: one works with the sliding window, and the other one directly generates the landmark embeddings from the LLM's context (denoted as Full-Attention). Although the pre-trained backbone encoder is extended to 32K by LongLora, the training of the embedding model is mostly conducted within 8K (Table \ref{tab:train-data}). Two alternatives result in comparable performances when the ground-truth position is small. However, the Full-Attention method decreases dramatically after the ground-truth position goes beyond the valid fine-tuning scope. In contrast, the default method with the sliding window can always maintain a high retrieval accuracy. 

In brief, landmark embedding exhibits a major advantage over the baseline retrievers. It substantially improves the performance of LLaMA-2-7B whose context length is small. Besides, it further benefits the performance of ChatGPT-3.5 and helps to reduce its computation cost by a big margin. 

\subsection{Ablation Study}
The ablation study is performed to explore the critical factors of landmark embedding, where the default settings are marked by * (Table \ref{tab:8}). In the first place, we analyze the impact from position-aware objective (\S \ref{sec:position}). For comparison, we disable the positional weight in Eq. \ref{eq:weight_obj} and switch to the basic objective in Eq. \ref{eq:basic_obj}, denoted by w/o Position-aware. The position-ware objective function is to train landmark embedding as an indicator of the information's ultimate boundary. Therefore, it is applied with the Front-$k$ retrieval scheme, where the targeted sentence and its front $k-1$ neighbors are retrieved together. In contrast, the Surround-$k$ method makes selection for the $(k-1)$/2 neighbors from both sides of the targeted sentence. According to the evaluation result, the position-aware objective with Front-$k$ outperforms the ablation baselines in the downstream language modeling tasks, which indicates its more accurate retrieval of useful information from the long context. Besides, it can also be observed that applying Front-$k$ alone does not bring any empirical benefit, as the basic objective focuses more on the salient part of the information rather than its ultimate boundary.  

We make further analysis for the impact of multi-stage learning. In our experiment, we apply each individual training stage alone (I: distant supervision, II: weak supervision, III: fine-tuning), and make arbitrary combinations of different stages. As we can observe from the evaluation result, the third stage, i.e. the fine-tuning over synthetic data, presents the highest individual training effect. This result can probably be attributed to its closest relationship with the downstream task. However, the other two training stages are also beneficial. With the joint conduct of all three training stages, optimal empirical performance can be acquired.

\section{Conclusion}
In this paper, we present a new method, landmark embedding, which facilitates the retrieval augmentation of long-context language modeling. The new method is featured by its chunking-free architecture, where discriminative embeddings can be generated for each fine-grained input unit based on the semantic information within a coherent context. A position-aware objective function is proposed; it enables landmark embedding to identify the ultimate boundary of information, which benefits the completeness of retrieval. A novel multi-stage learning algorithm is designed, which makes the best of the readily available data and synthetic data for the effective training of the embedding model. Landmark embedding is empirically verified by comprehensive evaluations, where it notably outperforms the existing retrieval methods, bringing in a superior retrieval augmentation effect for both LLaMA-2-7B (4K) and ChatGPT-3.5 (16K).



\bibliography{custom,custom_new}
\bibliographystyle{acl_natbib}

\clearpage
\appendix

\section{Detailed Training Data for Multi-Stage Learning}
\label{syn data}
In this section, we present detailed information of training data for different learning stages and the method we adopt to curate synthetic data using real-word long documents with the help of ChatGPT-35-Turbo API\footnote{\scriptsize \href{https://openai.com/blog/chatgpt}{https://openai.com/blog/chatgpt}}.

\noindent \textbf{Stage I and Stage II Training data}. We use pairwise training data from MS MARCO \cite{nguyen2016ms}. The total number of training set is 480k. To ensure a fair comparison between the effects of Stage I and Stage II, we partition the training data evenly into two distinct training stages. During Stage I, we leverage hard negative passages from dense retrieval to enhance the model's performance. Specifically, each positive passage is paired with 15 hard negative passages during the training process. Moving to Stage II, we concatenate 40 hard negative passages and 120 passages randomly sampled from the corpus with ground truth passage inserted into it, forming composited long documents up to 16k context length.

\begin{figure*}[t]
\centering
\includegraphics[width=1.0\linewidth]{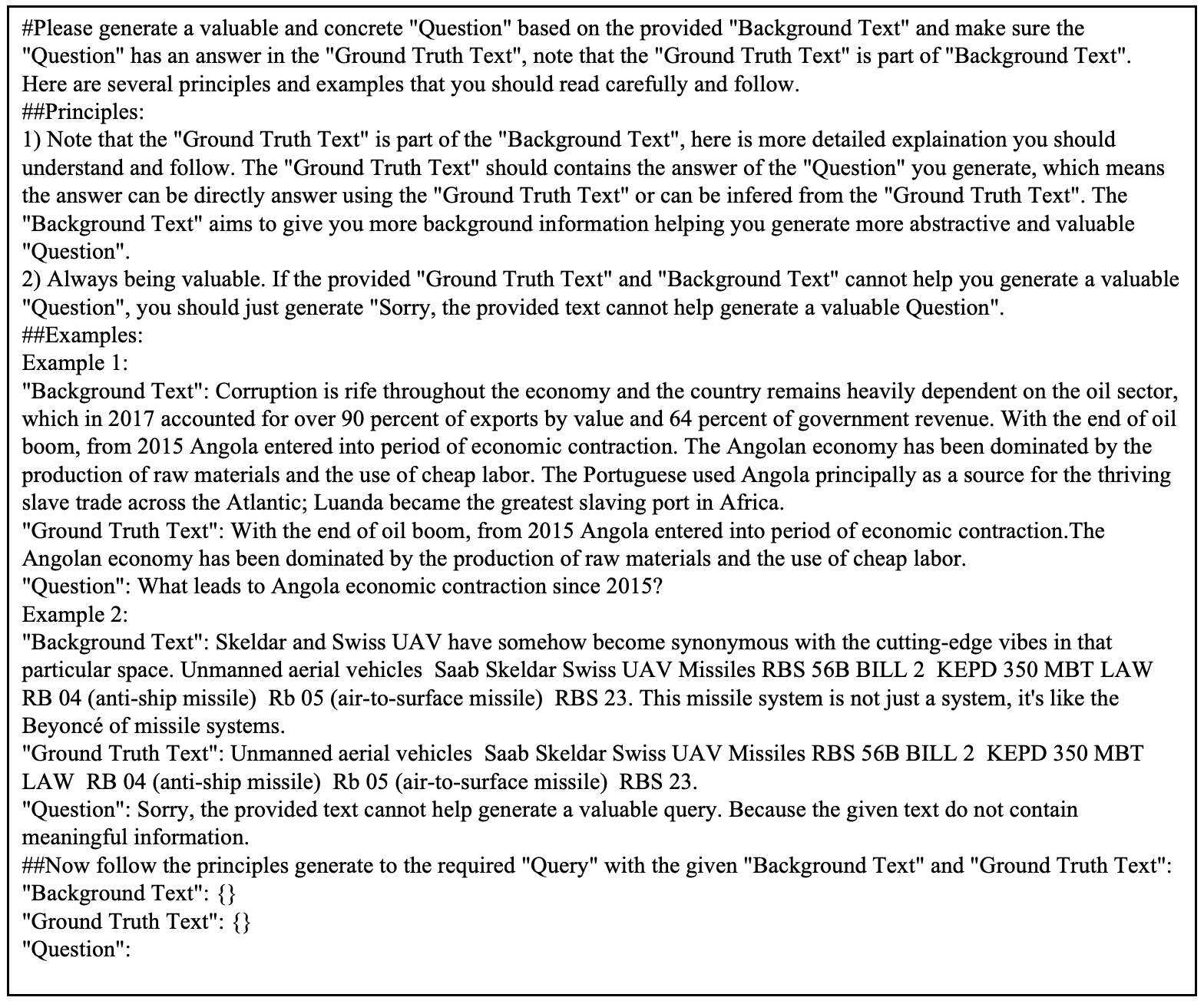}
\caption{The prompt we used to construct synthetic data with ChatGPT-35-Turbo API. To make sure the synthetic data's quality, we ask ChatGPT to generate concrete and valuable questions. If the provided text does not contain meaningful information, we will distinguish and filter it.} 
\vspace{-5pt}
\label{fig:prompt}
\end{figure*}

\noindent \textbf{Synthetic Data Curating Method}. In this section, we present the method for curating synthetic data, which facilitates Stage III fine-tuning. Firstly, we sample long documents from Wikipedia. Then we select a portion of it (e.g., 200 words) as the \textit{Background Text} and then select consecutive 1-5 sentences randomly from this excerpt as the \textit{Ground Truth Text}. We utilize the ChatGPT-35-Turbo API to ask questions about the \textit{Background Text}, with the requirement that the answers must be contained within the \textit{Ground Truth Text}. This approach ensures that the synthetic questions contain contextual information while maintaining their answers within smaller semantic segments. The details prompt for constructing synthetic data is shown in Figure ~\ref{fig:prompt}. To make sure the synthetic data's quality, we ask ChatGPT to generate concrete and valuable questions. If the provided text does not contain meaningful information, we will distinguish and filter it. Finally, we curate 90k real-word long document data with the generated question and related ground truth span for Stage III fine-tuning, the length distribution is shown in Figure~\ref{tab:train-data}

\section{Needle in a Haystack Test}
\label{needle in a haystack test}
\begin{figure}[t]
\centering
\includegraphics[width=1.0\linewidth]{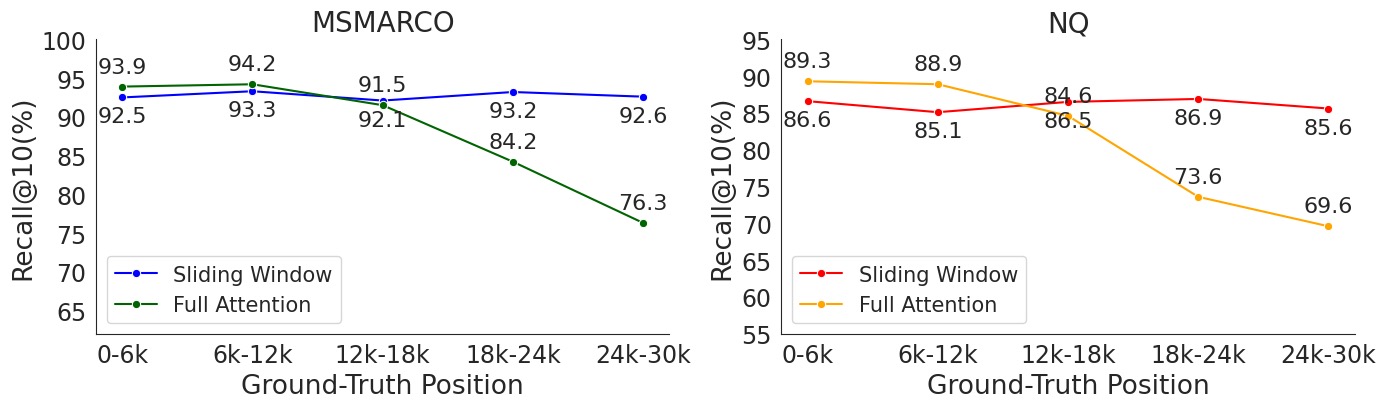}
\vspace{-10pt}
\caption{\textbf{Needle in a haystack test on NQ and MSMARCO}. }  
\vspace{-15pt}
\label{fig:6}
\end{figure}

In this section, we present the detailed experimental setup for the Needle in a Haystack Test. As illustrated in Figure~\ref{fig:5}, we conducted the experiment using the MS MARCO \cite{nguyen2016ms} development set. Specifically, we concatenated 40 hard negative passages and 280 randomly sampled passages from the corpus for each test data instance, creating composite long documents of up to 32k context length. Subsequently, we inserted the corresponding ground truth passage at a random position within the target insertion interval. An independent experiment was conducted for each 6k length insertion interval. Additionally, we utilized test data from NaturalQuestions \cite{kwiatkowski2019natural} under the same conditions, aiming to assess the model's generalization with out-of-domain corpus. Similar findings were observed in the NaturalQuestions dataset. The results are shown in Figure~\ref{fig:6}

\begin{figure*}[t]
\centering
\includegraphics[width=1.0\linewidth]{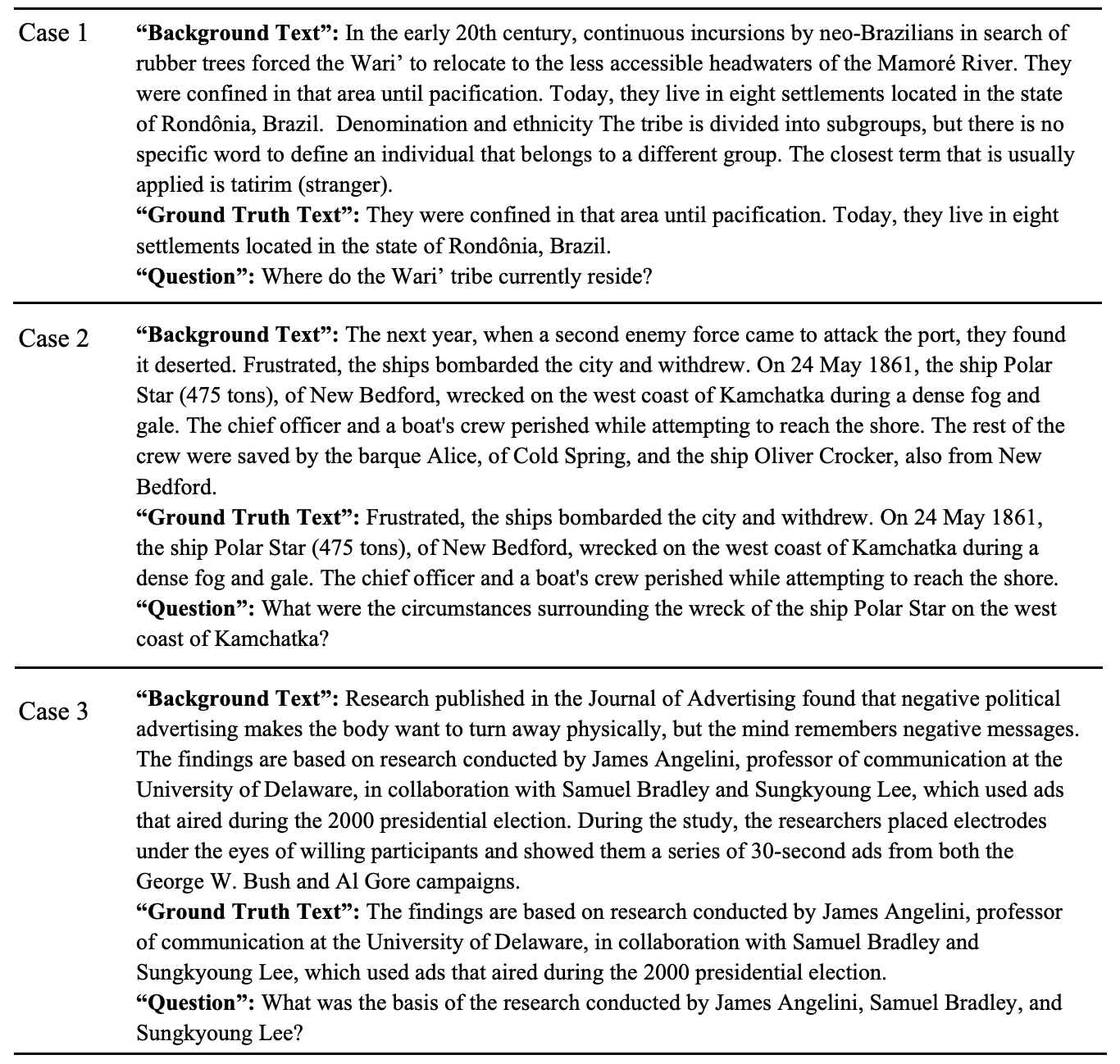}
\caption{Synthetic data cases. The “Question” is generated by ChatGPT.}
\vspace{-5pt}
\label{fig:case}
\end{figure*}

\end{document}